\documentclass[12pt]{article}
\usepackage{times}  
\usepackage{helvet} 
\usepackage{courier}  
\usepackage{graphicx} 
\usepackage{amssymb}
\usepackage{subfigure}
\usepackage{array}
\usepackage{multirow}
\usepackage{placeins}
\usepackage{natbib}
\usepackage{graphicx}  
\usepackage{url}
\usepackage[hidelinks]{hyperref}
\usepackage{graphicx}
\usepackage{amsmath}
\usepackage{amsthm}
\usepackage{algorithm}
\usepackage{algorithmic}
\usepackage{bm}

\usepackage{amsmath}
\usepackage{booktabs}
\usepackage{algorithm}
\usepackage{amssymb}
\usepackage{array}
\usepackage{multirow}
\usepackage{placeins}
\usepackage{natbib}
\usepackage{algorithmic}
\renewcommand{\cite}{\citep}

\usepackage[ruled,noresetcount,algo2e]{algorithm2e}
\newenvironment{sciabstract}{%
\begin{quote} \baselineskip14pt\small\hfil {\bf Abstract} \hfil\\[3pt]}
{\end{quote}\vspace{6pt}}

\newcounter{lastnote}

\title{Neural-to-Tree Policy Distillation with Policy Improvement Criterion}

\author{
Zhao-Hua Li$^1$, Yang Yu$^{1,2,\diamond}$, Yingfeng Chen$^{3}$, Ke Chen$^3$, Zhipeng Hu$^{3,4}$, Changjie Fan$^3$\\
\normalsize{
$^1$National Key Laboratory for Novel Software Technology, Nanjing University, China}\\
\normalsize{$^2$Polixir Technologies, China}\\
\normalsize{$^3$NetEase Fuxi AI Lab, China}\\
\normalsize{$^4$Zhejiang University, China}\\
\normalsize{$^\diamond$Corresponding author: yuy@nju.edu.cn}
}

\date{}

\begin{document}

\baselineskip16pt

\maketitle 

\begin{sciabstract}
While deep reinforcement learning has achieved promising results in challenging decision-making tasks, the main bones of its success --- deep neural networks are mostly black-boxes. A feasible way to gain insight into a black-box model is to distill it into an interpretable model such as a decision tree, which consists of if-then rules and is easy to grasp and be verified. However, the traditional model distillation is usually a supervised learning task under a stationary data distribution assumption, which is violated in reinforcement learning. Therefore, a typical policy distillation that clones model behaviors with even a small error could bring a data distribution shift, resulting in an unsatisfied distilled policy model with low fidelity or low performance. In this paper, we propose to address this issue by changing the distillation objective from behavior cloning to maximizing an advantage evaluation. The novel distillation objective maximizes an approximated cumulative reward and focuses more on disastrous behaviors in critical states, which controls the data shift effect. We evaluate our method on several Gym tasks, a commercial fight game, and a self-driving car simulator. The empirical results show that the proposed method can preserve a higher cumulative reward than behavior cloning and learn a more consistent policy to the original one. Moreover, by examining the extracted rules from the distilled decision trees, we demonstrate that the proposed method delivers reasonable and robust decisions.
\end{sciabstract}

\section{Introduction}
Deep reinforcement learning has achieved abundant accomplishments in many fields, including but not limited to order dispatching \cite{tang2019deep}, autonomous vehicles \cite{o2018scalable}, biology \cite{neftci2019reinforcement} and medicine \cite{popova2018deep}. However, its success usually depends on the Deep Neural Network (DNN), which has some pitfalls, for example, hard to verify \cite{bastani2016measuring}, lacking interpretability \cite{Deep_learning}, and so on. In contrast, the decision tree establishes an intelligible relationship between the features and the predictions, which can be also translated into rules to realize the robustness verification \cite{verif-tree}, and applied in security-critical areas \cite{healthcare}. However, decision trees are challenging to learn even in the supervised setting; there has been work that trains tree-based policy on dynamic environments \cite{ernst2005tree,liu2018toward}, but these algorithms have their limitations on generalizations (only scale to toy environments like \textit{CartPole}). To combine the advantages of both kinds of representation, a number of approaches based on knowledge distillation have been proposed \cite{frosst2017distilling,bastani2018verifiable}, i.e., neural-to-tree model distillation. Specifically, they trained a teacher DNN model, then used it to generate labeled samples, and finally fit the samples to a student tree, aiming to achieve robustness verification or model interpretation. 

In sequential decision-making tasks, the basic assumption in model distillation by supervised learning --- stationary data distribution is violated \cite{ross2010efficient}. If greedily imitating the teacher behavior, the student could drift away from the demonstrated states due to error accumulation, especially in complex environments \cite{ross2011reduction}. To address this issue, Viper uses the Dagger \cite{ross2011reduction} imitation learning approach to constantly collect state actions pairs for iteratively revising the student DT policies. The augmented data facilitates the learning, but could also cause inconsistency with the teacher --- some of which might never be encountered by the teacher but affect the tree's structure. Therefore, it is meaningful and valuable to explore how to effectively distill with the offline data.

In this paper, we propose a novel distillation algorithm, Dpic, which optimizes a novel distillation objective derived from the policy improvement criterion. Intuitively, the data-driven distillation schema has the same prerequisite as policy optimization --- a known policy model and some trajectories sampled by it. Different from Behavior Cloning, which minimizes the 0-1 loss, or Viper, which considers the long-term cost in the online setting, policy optimization, in particular, TRPO, proposes an improvement criterion over old policy on performance. The criterion can be used with offline data and inherently controls the distribution shift problem. However, the gradient-based optimization techniques are hard to apply to the nondifferentiable tree-based model. Moreover, explicit actions are needed when evaluating the improvement --- it is not enough to distinguish between the right actions from the wrong ones. We instantiate our approach by modifying the traditional decision tree learning algorithm --- \textit{info gain} for classification and \textit{CART} for regression, to optimize the policy improvement criterion; we show the new optimization objective can effectively improve the performance in offline data. Then, we integrate our objective with the offline Viper; we find that our algorithm can provide a more consistent policy without sacrificing the efficiency and model complexity. 

Overall, our main contributions in this work are: 
\begin{itemize}
    \item We proposed an advantage-guided neural-to-tree distillation approach, and analysed the connection between the advantage cost and the accumulative rewards of the distilled tree policy.
    \item Two practical neural-to-tree distillation algorithms are devised and tested in several environments, showing that our methods performs well in both the terms of the average return and the consistency of state distribution.
    \item We investigate the interpretability of the distilled tree policies in the fighting game and the self-driving task, showing that our distilled decision tree policies deliver reasonable explanations.
\end{itemize}

\section{Related Work}
\subsection{Policy Distillation}
Unlike other model distillation methods for RL \cite{policy_distillation,dis_policy_distillation}, which aim to satisfy the computational demand or timing demand, our work falls into the cross-model transfer --- neural-to-tree policy distillation. Some work \cite{coppens2019distilling} utilizes the distillation to explain existing policy but only limits it in the traditional supervised methods. The most related work to ours is Viper \cite{bastani2018verifiable}, which uses the DAgger \cite{ross2011reduction}, to constantly revise the policy in an online way and also takes into account the Q-function of teacher policy to prioritize states of critical importance. In contrast, the proposed method, derived from the policy improvement, evaluates, and maximizes the advantage of the distilled policy. As a result, the decision-tree learning in Viper only discriminates against the right actions from the wrong actions, while the proposed method evaluates every action and tries to select the best one. Moreover, our tree-based policy learning algorithm can achieve good enough performance offline to avoid model distortion due to data aggregation. It can also be regarded as an extension of cost-sensitive learning.  

\subsection{Model-Agnostic Interpretation Method}
Our work can be regarded as a tool to implement the model-agnostic interpretation \cite{doshi2017towards,molnar2019interpretable}, which separates the explanations from the machine learning model. There has been work in the domain, such as providing feature importance \cite{NIPS2019_9211} that influences the outcome of the predictor and generating counterfactual inputs that would change the black box classification decision \cite{chang2018explaining,singla2019explanation}, which all provide a visual explanation of the decision. These methods largely focus on local explanations or analyzing the key area to influence the prediction. Another research direction is model distillation, which distills knowledge from the black-box model into a structured model to implement global interpretability \cite{frosst2017distilling,zhang2019interpreting}. As an effective neural-to-tree distillation method, our work belongs to the latter, aiming to make the entire RL policy model transparent. Due to the structured model, this method can meet verification, safety, and non-discrimination easily.

\section{Distilling Deep Models into Tree Models}
Deep neural network models encode their knowledge in the connections between neurons. Benefiting from these excellent fitting and generalization abilities, deep models are quite flexible and powerful in complex learning tasks. However, the distributed knowledge representation blocks their interpretability, which could pose potential dangers in real applications. Decision trees have been widely used in data mining and machine learning as a comprehensible knowledge representation, which can be easily comprehended and efficiently verified. Therefore, transforming a deep model into a decision tree model is a feasible way to break the black-box barrier of DNN. In this section, we firstly present a general decision tree learning process and then introduce previous studies on distillation into decision trees.

\subsection{Decision Tree Learning}
Decision tree learning, as a predictive modeling approach, is commonly used in machine learning and data mining. A tree is a hierarchy of filters in its internal nodes to assign input data to a leaf node with prediction.
By ``testing'' on the features continuously, i.e., adding if-then rules, the decision tree divides up the feature space, using the information from data to predict the corresponding value or label.
Formally, the decision tree (commonly binary tree) $\mathbb T$ is built by dividing the training data $\mathcal{D}$ into subsets $\mathcal{D}_L$ and $\mathcal{D}_R$ recursively. 
The split point $x\in\mathcal{X}$ for each node $\mathbb{N}$ is chosen by a certain criterion, and the splitting will repeat recursively until the termination condition meets.

\subsection{Node Splitting Criterion}

\subsubsection{Error reduction}
Denote the misclassification error as $E$, data size as $N$, and the misclassification error rate as $\mathcal E=E/N$. A common target of decision tree learning is to minimize the sum of the misclassification error $E$ at each leaf node $\mathbb N$: 
\begin{small}$\sum_{\mathbb N=1}^{|\mathbb T|}E_{\mathbb N}(\mathbb T)$\end{small}.
In the process of tree construction, a straightforward strategy is that the split point $x$ in each $\mathbb N$ is chosen by maximizing the \textit{error reduction}:
\begin{equation}
    R^E_\mathbb N = \mathcal E(\mathcal D_\mathbb N)-\frac{N_{\mathbb N,L}}{N_\mathbb N}\mathcal E(\mathcal D_{\mathbb N,L}|x)-\frac{N_{\mathbb N,R}}{N_\mathbb N}\mathcal E(\mathcal D_{\mathbb N,R}|x)
\end{equation}
where the subscripts $\mathbb N, L$ and $\mathbb N, R$ represent  $\mathbb N$'s child-nodes.

\subsubsection{Cost reduction}
Here we consider misclassification cost instead of 0-1 error.
Denote the cost of a sample in terms of classification to the label $k \in \mathcal Y$ as $C^k$, the cost labeled $k$ in $\mathbb N$ as 
\begin{small}$C_{\mathbb N}^k=\sum_{\mathcal{D}_\mathbb N}C^k$\end{small}
, and the total cost at each leaf node $\mathbb N$ as  
\begin{small}$W_\mathbb N = \sum_{k\in\mathcal{Y}}C^k_\mathbb N$ \end{small}.
Aligning with the traditional greedy algorithm, we define the classification cost rate \begin{small}
$\mathcal C^k={C^k}/W$
\end{small} and the minimal classification cost rate 
\begin{small}
    $\mathcal C=\min_{k\in\mathcal{Y}}\mathcal C^k = \min_{k\in\mathcal Y}{C^k}/W$
\end{small}.
Now the objective of decision tree is:
\begin{small}
    $\sum_{\mathbb N=1}^{|\mathbb T|}C_\mathbb N(\mathbb T)$
\end{small}.
Correspondingly, the greedy node-splitting strategy is to maximize the \textit{cost reduction}:
\begin{equation}
    R^C_\mathbb N = \mathcal C(\mathcal D_\mathbb N)-\frac{W_{\mathbb N, L}}{W_\mathbb N}\mathcal C(\mathcal D_{\mathbb N,L}|x)-\frac{W_{\mathbb N, R}}{W_\mathbb N}\mathcal C(\mathcal D_{\mathbb N, R}|x)
\end{equation}

\subsubsection{Node-splitting degradation problem}
From the definition, we can find that $\forall k \in \mathcal Y$,
\begin{equation}
    \mathcal C^k(\mathcal D_\mathbb N)=\frac{W_{L}}{W}\mathcal C^k(\mathcal D_L)+\frac{W_R}{W}\mathcal C^k(\mathcal D_R)
\end{equation}
It means that if the labels of child nodes split by each split point are identical with the parent node, $R^C$ will be equal to zero, and the split point choice will be arbitrary. Formally, if 
\begin{small}
$\forall x\in \mathcal X, \arg_{k \in\mathcal Y}\mathcal C(\mathcal D)=\arg_{k \in\mathcal Y}\mathcal C(\mathcal D_L|x)=\arg_{k \in\mathcal Y}\mathcal C(\mathcal D_R|x)$,
\end{small}
then 
$
    R^C\equiv 0
$.
We call it \textit{node-splitting degradation problem}. The mathematical cause is that the function $f(\mathbf{z})=\min\{z_i\}$ is not strictly convex. That is, if 
\begin{small}
${\exists} \mathbf{z'}, \mathbf{z''}, \arg\min \{z'_i\}=\arg\min \{z''_i\}=\arg\min \{\theta z'_i + (1-\theta)z''_i\}$,
\end{small}
then
\begin{equation}
    f(\theta \mathbf z' + (1-\theta)\mathbf z'') = \theta f(\mathbf z') + (1-\theta)f(\mathbf z'')
\end{equation}
Define $
\mathcal O = \{(\mathbf z', \mathbf z'',  \theta \mathbf z' + ( 1- \theta ) \mathbf z'' ) | \arg \min \{z'_i\} = \arg \min \{ z''_i \} = \arg \min \{\theta z'_i + (1-\theta)z''_i\}  \}$.
If $\forall x\in \mathcal X$, $(\mathcal C(\mathcal D_{L}|x), \mathcal C(\mathcal D_R|x), \mathcal C(\mathcal D)) \in \mathcal O$, then the evaluation for each split point will be identical, which could lead to a bad tree.

\subsubsection{Entropy form}
To tackle the problem, previous studies introduced the entropy related evaluations like \textit{info gain}.
The corresponding information gain is 
\begin{equation}
    IG^\mathcal C_\mathbb N = H_{\mathcal C}(\mathcal D_\mathbb N) - \frac{W_{\mathbb N,L}}{W_\mathbb N}H_{\mathcal C} (\mathcal D_{\mathbb N,L}|x) - \frac{W_{\mathbb N,R}}{W_\mathbb N}H_{\mathcal C}(\mathcal D_{\mathbb N,R}|x)
\end{equation}
where $H_{\mathcal C}(\mathcal D)$ indicates the entropy of cost distribution on $\mathcal D$, i.e., $$H_{\mathcal C}(\mathcal D)=-\sum_{k \in\mathcal Y}\mathcal C^k(\mathcal D)\log \mathcal C^k(\mathcal D).$$
 
Compared to the min function $\mathcal C(\mathcal D)=\min_{k \in \mathcal Y}\{\mathcal C^k(\mathcal D)\}$, the entropy form $H_{\mathcal C}(\mathcal D)$ has three properties that 1) is strictly convex;  2) has the same saddle points; 3) has the same monotonicity for each variable. These properties ensure that the objective can avoid the problem of degradation without effecting the optimization results. The detailed proof is attached in Appendix A.

\subsection{Distilling Deep Model Agents into Trees}
In the general formal of distillation, knowledge is transferred to the distilled model by training it on a transfer set. In RL, agents make decisions according to the decision models or the predictive models, i.e., the knowledge is usually stored in the decision behavior or the predictions for the future.
Correspondingly, there are two approaches to achieve knowledge distillation from an existing agent by constructing different transfer sets --- decision data-driven and predictive data-driven \cite{policy_distillation}. The former implements the distillation by imitating the demonstrated action greedily, i.e., behavior cloning. The latter approximates the Q-function of the DNN policy by a regression tree and afterwards chooses the action with highest ``quality''. Both transfer data are generated and collected by the policy to be distilled interacting with the environment.

\begin{table*}[t]
    \renewcommand\arraystretch{1.2}
  \centering
  \begin{tabular}{lll}
  \toprule
    Name    &    Objective function    &     Distribution to be evaluated \\
   \midrule
    BC    & 
    $\sum_{s\in \mathcal{D}}\mathbb{I}(\mathbb T(s)\neq \pi^{*}(s))$
    &
    $E_{\mathbb N}^a=\sum_{s\in \mathcal D_{\mathbb N}} \mathbb{I}(a\neq \pi ^ \text{n}(s)) $
    \\
    Dpic
    &
    $\sum_{s\in \mathcal{D}}-A(s, \mathbb T(s))$
    &
    $C_{\mathbb N}^a=\sum_{s\in \mathcal D_{\mathbb N}}-A(s,a)$
    \\
    Dpic$^R$
    &
    $
    \sum_{s\in \mathcal{D}}-A(s, \mathbb T(s))    +\alpha \mathbb I(\mathbb T(s)\neq \pi^{*}(s))
    $
    &
    $
    C^{\text{R, a}}_{\mathbb N}=\sum_{s\in \mathcal D_{\mathbb N}}-A(s,a)
    +\alpha \mathbb I(a \neq \pi^{*}(s))
    $
    \\
   \bottomrule
  \end{tabular}
  \caption{The algorithms}
  \label{comparison-table}
\end{table*}

\section{Distillation with Policy Improvement Criterion}
In this section, we discuss the problem that the traditional algorithm (behavior cloning) faces in distillation and then propose our distillation objective.
\subsection{Distribution Shift Problem}
\label{section distributin shift}
In supervised learning, the i.i.d. assumption is often made for both training data and testing data to imply that all samples stem from the same probability distribution. But in reinforcement learning, the state distribution is dependent on the policy, which means that an inappropriate behavior not only misses instant reward but also causes the data distribution shift, leading to an extra loss in cumulative rewards.
\subsubsection{Error accumulation}
Consider a student policy $\hat \pi$ that imitates an expert policy $\pi^*$ by behavior cloning, the training data (transfer set) is under the distribution induced by the expert: $d_{\pi^*}$ while the testing data is under $d_{\hat \pi}$.
Denote the immediate cost of doing action $a$ in state $s$ as $l(s, a)$, its expectation under the policy $\pi$ as $l_{\pi}(s)=\mathbb{E}_{a\sim\pi_s}(l(s,a))$, the expected $T$-step cost as $J(\pi)=T\mathbb{E}_{s\sim d_{\pi}}(l_\pi(s))$, and the 0-1 loss of executing action $a$ in state $s$ as $e(s,a)$. According to \cite{ross2010efficient}, if $\mathbb{E}_{s\sim d_{\pi}^*}[e_\pi(s)]\leq \epsilon$, then
\begin{equation}
\label{error-accum-1}
    J(\hat{\pi})\leq J(\pi^*)+T^2\epsilon
\end{equation}
But if we have the long-term cost $l^l_\pi(s)=Tl_\pi(s)$ which satisfies 
$\mathbb{E}_{s\sim d_{\pi}^*}[l^l_\pi(s)]\leq \epsilon^l$, then 
\begin{equation}
\label{error-accum-2}
    J(\hat{\pi})\leq J(\pi^*)+T\epsilon^l
\end{equation}


\subsection{Policy Improvement Criterion}
Define an infinite-horizon discounted Markov decision process by the tuple $(\mathcal{S}, \mathcal{A}, P, r, \rho_0, \gamma)$ with state space $\mathcal{S}$, action $\mathcal{A}$, state transition probability distribution $P: \mathcal S \times \mathcal A\times\mathcal S\rightarrow [0, 1]$, rewards function $r: \mathcal S \times \mathcal A \rightarrow R$ , initial state distribution $\rho_0 $ and the discounted factor $\gamma$.  Given a policy $\pi$, its expected discounted reward $\eta(\pi)$:
\begin{equation}
    \eta(\pi) = \mathbb{E}_{s_0, a_0,...\sim (\rho_0,\pi, P)}\left[\sum_{t=0}^{\infty}\gamma^tr_t(s_t)\right]
\end{equation}
where the notation $s_0, a_0, s_1...\sim (\rho_0,\pi, P)$ indicates $s_0\sim\rho_0(\cdot)$,$a_t\sim\pi(\cdot|s_t)$, $s_{t+1}\sim P(\cdot|s_t,a_t)$.

Recall the standard definitions of the state-action value function $Q_\pi$, the value function $V_\pi$, and the advantage function $A_\pi$ are:
\begin{equation}
Q_\pi(s_t, a_t)=\mathbb{E}_{s_{t+1},a_{t+1},\ldots\sim(\pi, P)}\left[\sum_{l=0}^{\infty}\gamma^k r(s_{t+l})\right]
\end{equation}
\begin{equation}
    V_\pi(s_t)=\mathbb{E}_{a_t, s_{t+1},\ldots\sim(\pi, P)}\left[\sum_{l=0}^{\infty}\gamma^k r(s_{t+l})\right]
\end{equation}
\begin{equation}
    A_\pi(s, a)=Q_\pi(s,a)-V_\pi(s)
    \label{advantage}
\end{equation}
Then the expected return of student policy $\hat \pi$ can be expressed in terms of the advantage over expert policy $\pi^*$, accumulated over timesteps (the detailed proof is analogous to \cite{schulman2015trust}):
\begin{equation}
    \label{equ: policy update 1}
    \eta(\hat \pi)=\eta(\pi^*)+\mathbb{E}_{s_0, a_0,...\sim (\rho_0,\hat \pi, P)}\left[\sum_{t=0}^{\infty}\gamma^kA_{\pi^*}(s_t, a_t)\right]
\end{equation}
Rewrite the Equation (\ref{equ: policy update 1}) in terms of the sum over states $\rho_\pi(s)=\sum_{t=0}^{\infty}\gamma^tP(s_t=s)$, we have:
\begin{equation}
    \label{equ: policy update 2}
    \eta(\hat \pi)= \eta(\pi^*) + \sum_s\rho_{\hat\pi}(s)\sum_a\hat\pi(a|s)A_{\pi^*}(s,a),
\end{equation}
which means when we sample a batch of data from $\rho_{\hat\pi}$, we can use the $\pi^*$'s information (the advantage $A_{\pi^*}(s,a)$) to have an assessment for $\hat \pi$'s accumulated rewards.

\subsubsection{Distillation objective}
Now we use a decision tree $\mathbb T$ to mimic the DNN policy $\pi^*$, and the loss can be defined as
\begin{equation}
\begin{split}
    \ell(\mathbb T)=\eta(\pi^*)-\eta(\mathbb T) 
    &= -\sum_s\rho_{\mathbb T}(s)A_{\pi^*}(s,\mathbb T(s))\\
    &= -\sum_{s\in \mathcal{D_{\mathbb T}}}A_{\pi^*}(s, \mathbb T(s)) 
\end{split}
\end{equation}
Here $\mathcal D_\mathbb T$ indicate the data is sampled using the policy $\mathbb T$.
Considering that the offline data $\mathcal D_{\pi^*}$ is collected by $\pi^*$, we can rewrite the objective:
\begin{equation}
    \ell(\mathbb T) = \sum_{s\in \mathcal{D_{\pi^*}}}-\frac{\rho_{\mathbb T}(s)}{\rho_{\pi^*}(s)}A_{\pi^\text{n}}(s, \mathbb T(s))
\end{equation}
When transfer data is infinite and the distilled tree has no complexity limitation, the generated policy $\mathbb T$ will be extremely similar to the existed policy $\pi^*$, and the difference of corresponding state distributions could be negligible. An objective can be given:
\begin{equation}
\label{objective1}
    \hat{\ell}(\mathbb T) = \sum_{s\in \mathcal{D_{\pi^*}}}-A_{\pi^\text{n}}(s, \mathbb T(s))
\end{equation}
Now we can use the advantage information (characterizes the long-term effects in the offline setting) as a classification cost to construct the tree. 

\subsubsection{Regularization}
As mentioned before, maximizing the pure advantage ignores the distribution mismatch problem caused by policy differences so that it only works when the tree is large. The main reason is that the new objective inherently distinguishes samples according to the advantage information, which hinders the tree's fitting ability. In other words, to fit identical data, optimizing the new objective generates a larger tree than behavior cloning. However, it is usually necessary to limit the model complexity when we refer to the model explanation, in which case the distribution mismatch problem is usually non-negligible. To lessen this problem, we further consider the objective of behavior cloning as a penalty term. Now we have the second objective:
\begin{equation}
    \hat\ell^{R}(\mathbb T)=\sum_{s\in\mathcal D_{\pi^{*}}}-A_{\pi^*}(s, \mathbb T(s))+\alpha \mathbb I (\mathbb T(s)\neq \pi^{*}(s))
\end{equation}
where the temperature coefficient $\alpha$ determines the strength of the penalty and is set prior to training. With $\alpha$ increasing, the affect of advantage information on tree will fade away. 
The algorithms are listed in Table \ref{comparison-table}.

\subsubsection{Advantage evaluation}
The calculation of the advantage requires $Q_{\pi^*}$ and $V_{\pi^*}$, but only one of them can be obtained in most RL algorithms. For policy gradient methods, the value function and action distribution are available, while for Q-learning based methods, the state-action value is accessible.
Inspired by the prior work --- SQIL \cite{reddy2019sqil}, we assume that the behavior of the DNN agent follows the maximum entropy model \cite{levine2018reinforcement}, and derive the advantage based on soft Q-learning \cite{haarnoja2017reinforcement}. 
The DNN policy $\pi^*$ and the value function $V_{\pi^*}$ can be defined by $Q_{\pi^*}$ :
\begin{equation}
    \label{softq}
    \pi^*(a|s)=\frac{\exp{(Q_{\pi^*}(s,a))}}{\sum_{a'\in \mathcal A} \exp{(Q_{\pi^*}(s, a'))}}
\end{equation}
\begin{equation}
    V_{\pi^*}(s)=\log{\sum_{a\in\mathcal A}
    \exp{( Q_{\pi^*}(s,a)})}
\end{equation}
Therefore, we can convert the form between $Q_{\pi^*}$ and $V_{\pi^*}$. That is, our algorithm can be applied to all neural network policies without considering how these policies are trained.

\begin{table}[t]
    \centering
            \begin{tabular}{c|ccccccc}
            \toprule
            n &BC &FQ &Viper$_M$ & Dpic &Dpic$_M$ &Dpic$^R$ &Dpic$^R_M$ \\
            \midrule
            \multicolumn{8}{c}{CartPole} \\
            \midrule
            $1$&$\bm{70.4}$&$9.5$&$28.9$&$14.8$&$32.1$&$56.7$&$38.7$\\
            $3$&$175.7$&$9.3$&$171.7$&$17.8$&$166.8$&$175.7$&$\bm{189.7}$\\
            $7$&$167.8$&$12.1$&$181.2$&$112.5$&$154.3$&$\bm{194.0}$&$186.1$\\
            $15$&$176.9$&$12.5$&$179.2$&$168.3$&$159.7$&$\bm{192.0}$&$187.4$\\
            $31$&$190.0$&$26.4$&$168.6$&$\bm{199.2}$&$181.9$&$194.7$&$185.3$\\
            $63$&$\bm{190.8}$&$25.9$&$167.5$&$171.1$&$181.2$&$186.9$&$190.5$\\
            \midrule
            \multicolumn{8}{c}{MountainCar} \\
            \midrule
            $1$&$-117.4$&$-200.0$&$-121.3$&$-117.4$&$-139.9$&$\bm{-116.5}$&$-120.0$\\
            $3$&$-118.3$&$-200.0$&$-118.3$&$-118.7$&$-117.0$&$-118.4$&$\bm{-110.0}$\\
            $7$&$-107.7$&$-200.0$&$-110.0$&$-105.8$&$-114.0$&$\bm{-105.3}$&$-105.3$\\
            $15$&$-111.1$&$-200.0$&$\bm{-104.3}$&$-106.9$&$-116.2$&$-105.3$&$-105.3$\\
            $31$&$-105.5$&$-200.0$&$-105.5$&$-106.0$&$-102.5$&$-100.0$&$\bm{-99.3}$\\
            $63$&$-95.0$&$-200.0$&$-105.3$&$\bm{-94.2}$&$-105.5$&$-96.6$&$-97.5$\\
            \midrule
            \multicolumn{8}{c}{Acrobot} \\
            \midrule
            $1$&$-84.3$&$-500.0$&$-97.0$&$-86.9$&$-94.0$&$-79.4$&$\bm{-79.0}$\\
            $3$&$-85.7$&$-500.0$&$-92.6$&$-83.5$&$-95.8$&$\bm{-82.5}$&$-98.8$\\
            $7$&$-82.4$&$-500.0$&$-88.5$&$-94.8$&$-130.8$&$-86.9$&$\bm{-82.0}$\\
            $15$&$-86.7$&$-500.0$&$-99.2$&$-83.0$&$-140.4$&$-89.3$&$\bm{-81.7}$\\
            $31$&$-93.6$&$-500.0$&$-98.5$&$-90.0$&$-105.0$&$\bm{-82.5}$&$-86.2$\\
            $63$&$-85.1$&$-500.0$&$-89.2$&$-86.6$&$\bm{-78.6}$&$-82.2$&$-81.5$\\
            \bottomrule
        \end{tabular}
    \caption{The average returns of distill trees across ten runs.}
    \label{table-offline}
\end{table}

\section{Evaluation}
In this section, we evaluate the proposed algorithms on four Gym tasks, including three classic control tasks and a variant of Pong game \cite{bastani2018verifiable} in Atari, which extracts 7-dimensional states from image observation. The DNN policies on classic control tasks are trained by Ape-X \cite{horgan2018distributed}, and for Pong, we adopt the optimal model provided in \cite{bastani2018verifiable}. The mean cumulative rewards obtained by the teachers on CartPole, MountainCar, Acrobot and Pong are $200.0$, $-63.2$, $-102.0$ and $21.0$, respectively. All results are averaged across 100 episodes. The algorithm with the regularization term has an additional hyper-parameter $\alpha$; we describe how we choose it below. Except for our algorithms, other algorithms adopt the standard decision tree algorithms (i.e., \textit{ID3} and \textit{CART}) in scikit-learn \cite{scikit-learn}.

\subsection{Comparison in Offline Setting}
On the classic control tasks, we compare our algorithms to existed other distillation methods, including behavior cloning (BC), fitting Q (FQ), and the offline Viper without data aggregation (Viper$_M$). The process of our tree construction is the same as the classical decision tree algorithm. In particular, for each tree node, we choose the split-point by minimizing $\hat{\ell}$ (Dpic) or $\hat{\ell}^R$ (Dpic$^R$) (summarized in Table \ref{comparison-table}). The transfer data (including the observation, the selected action, and the Q values of all actions) is collected by interacting with environments using well-trained neural network policies. Viper$_M$, Dpic$_M$, and Dpic$^R_M$ resample data as well as Viper, but others do not. In algorithms with regularization terms, we explore the following values for the error cost weighting coefficient $\alpha$, $\{0.02, 0.04, 0.08, 0.1, 0.15\}$ and choose the best one by the returns. 

We train decision trees with maximum node numbers of $\{1, 3, 7, 15, 31, 63\}$ and run ten times by different fixed data in each setting. As shown in Table \ref{table-offline}, our algorithms can obtain the highest rewards in $15/18$ settings, regardless of the size of the tree. Moreover, with the increase of the number of nodes and the enhancement of tree expression ability, the difference between teacher and student is decreasing, which leads to a better result of the simplified algorithm ($199.2$ in CartPole, $-94.2$ in MounatinCar, and $-78.6$ in Acrobot). Dpic and Dpic$^R$ can thus apply to policy distillation with any limitations on model complexity.

\begin{figure}[t]
    \centering
    \subfigure[average rewards]{
    \label{average rewards}
    \includegraphics[width=0.47\linewidth]{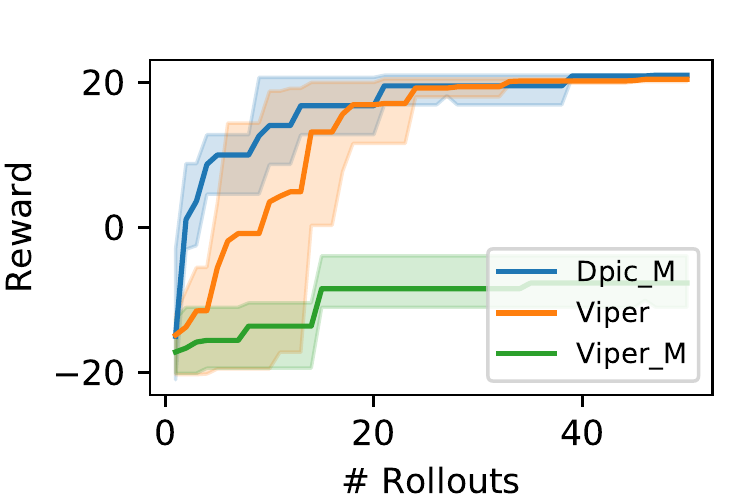}
    }
    \subfigure[state distribution discrepancy]{
    \label{Maximum Mean Depency}
    \includegraphics[width=0.47\linewidth]{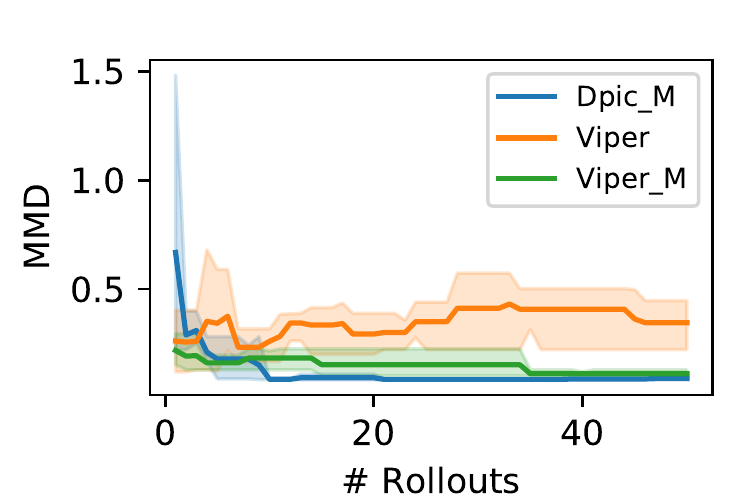}
    }
    \caption{Comparison to Viper in Atari Pong.
    }
    \label{pong}
\end{figure} 

\subsection{Comparison to VIPER}
On the Atari Pong benchmark, we integrate our algorithm to the offline Viper (always collect data using the teacher) and compare it to the standard and offline Viper. In Figure \ref{average rewards}, we compare the reward achieved by three algorithms as a function of the number of rollouts (the experiment runs three times by different seeds). Moreover, we also compare the state distribution discrepancy using the maximum mean discrepancy (\textit{MK-MMD} \cite{gretton2012kernel}), which is a distance measure on the probability space. All algorithms limit the tree-size to 80. Dpic achieves comparable performance and efficiency with Viper but a much better consistency ($0.0871$ VS $0.3445$ in MMD). In contrast, the Viper without data aggregation has a equivalent model fidelity but a worse performance ($-7.6$ VS $21.0$). 


\begin{figure*}[t]
    \centering
    \subfigure[skills tree]{
    \label{skills tree}
    \includegraphics[width=0.8\linewidth]{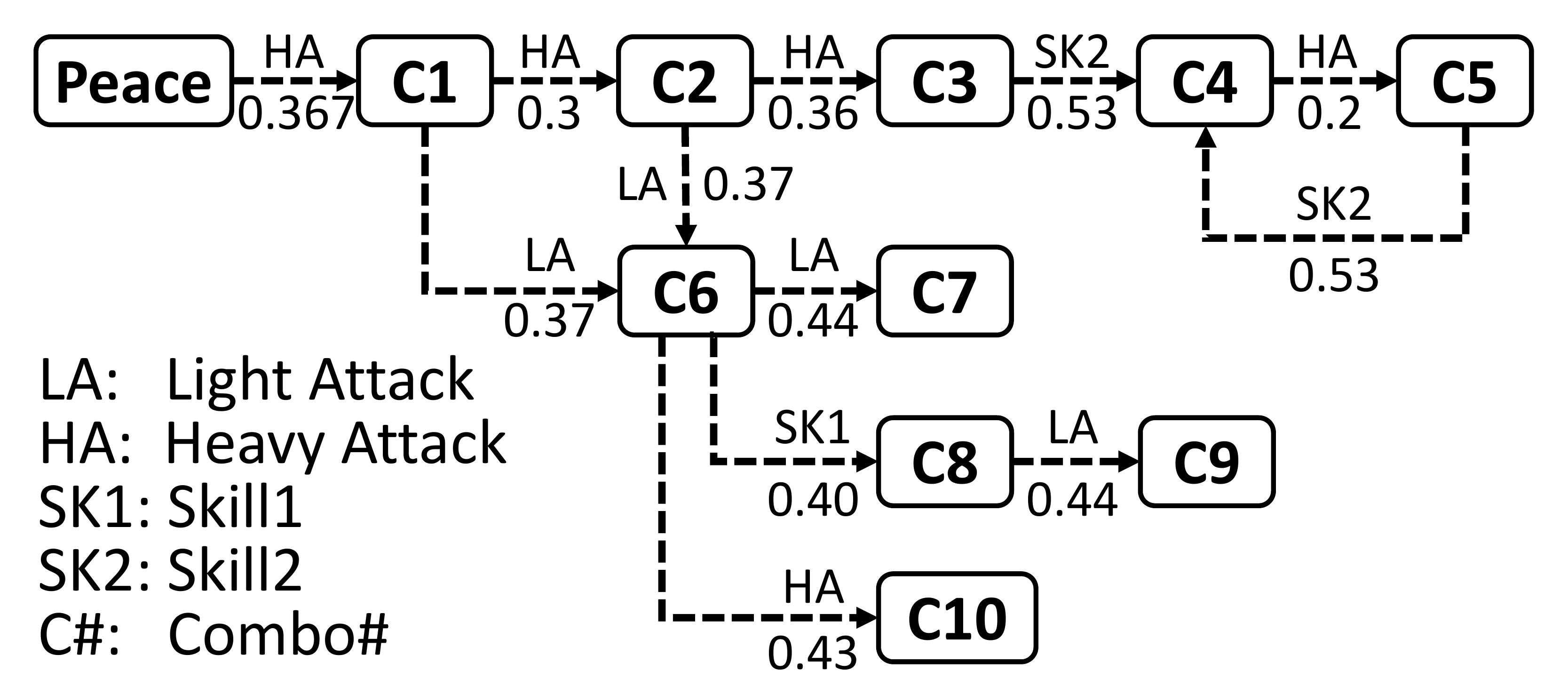}
    }
    \subfigure[feature importance]{
    \label{feature importance}
    \includegraphics[width=0.95\linewidth]{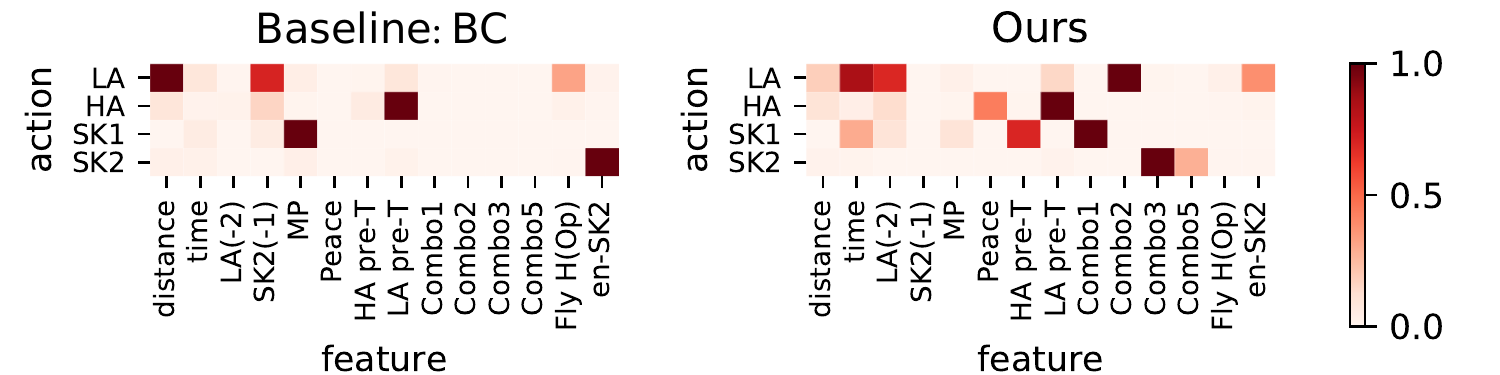}
    }
    \caption{The attack actions explanation on \textit{Fighting Game}. Fig \ref{skills tree} shows the inherent skill process in the game. We count the top-1 important feature which the model considers in ten thousand decisions and show the heat map in Fig \ref{feature importance}. 
    By comparing the feature of most concern when taking the attack actions, we can find the advantage-based method provide a more reasonable result --- paying more attention to the agent skill status (e.g., \textit{Combo} 1, 2, 3) instead of the basic information (distance, mana point).
    }
    \label{game}
\end{figure*} 

\section{Policy Explanation}
Here we mainly make some trials on policy explanation through feature importance in a commercial fighting game and an autopilot simulation environment. These tasks are more complex: the fighting game requires masterly combo attack to beat the opponent while the autopilot simulator needs to avoid pedestrians and arrive the destination. The state sizes of both tasks are about 200. Besides, we also illustrate the online performance of the distilled trees.
The results shown in Table \ref{big-experiments}, and our methods both learn a better-performing tree in two tasks.

\begin{table}[htp]
  \centering
  \begin{tabular}{lcc}
  \toprule
        &   Fighting Game    &  Autopilot Simulation  \\
     \midrule
    Random & $44.32 \pm 19.87$ & $8.4514 \pm 11.02$ \\
    Q-distillation  & $181.81 \pm 24.32$ & $14.7221 \pm 6.51$ \\
    BC  & $ 187.70 \pm 18.82 $&$15.4205 \pm 7.10$ \\
    Dpic  & $190.78 \pm 17.47$ & $15.5012 \pm 7.94$ \\
    Dpic$^R$  & \bm{$195.24 \pm 12.35$} & \bm{$16.0453 \pm 6.87$}\\
    Viper$_M$ & $ 189.31 \pm 19.08 $ & $ 15.5540 \pm 7.20 $ \\
    \midrule
    DNN(Expert) & $194.20 \pm 17.84$ & $16.5686 \pm 6.36$\\
  \bottomrule
  \end{tabular}
  \caption{The average returns of distilled trees.}
  \label{big-experiments}
\end{table} 

\subsubsection{Fighting Game}
The goal of the task is to beat the opponent in a limited time. The opponent can not fight back when suffering damage, so the best strategy is to perform consecutive strikes according to the skill tree (shown in Fig \ref{skills tree}) as much as possible.
To explore how the agent fights with the opponent, we analyse on the feature importance of four attack actions.
Traditional feature importance is an average rate that the decrease rate of impurity occurs on the splits between the root node and leaf node. 
We extend the node impurity to the cost ratio and the value indicates the importance of the feature likewise.
Figure \ref{feature importance} shows the statistical results of the top one important features.

From our result, we can find that the agent pays attention to the combo attack, e.g., the \textit{light attack} is related to the \textit{combo2} status, the \textit{skill1} and \textit{skill2} are respectively related to the \textit{combo1} and the \textit{combo3}. On these combo statuses, the corresponding actions will ensure the combo attack running smoothly referring to Figure \ref{skills tree}. By contrast, the baseline result demonstrates the importance of distance, mana points, and whether the skill is available, which are more basic.
We also find that the agent focuses on the preparation time (\textit{foreswing} in the game) of other actions during attack process. We argue that the agent infers the current combo statuses based on the time because the preparation time of behaviors in different combo status will change.
All in all, our algorithm, by focusing on the reward information, provide a more objective-oriented explanation; while BC pay more attention on the numbers of state, and the distilled tree has more statistical characters. 

\subsubsection{Autopilot Simulation}
\begin{figure}[t]
    \centering
    \includegraphics[width=0.8\linewidth]{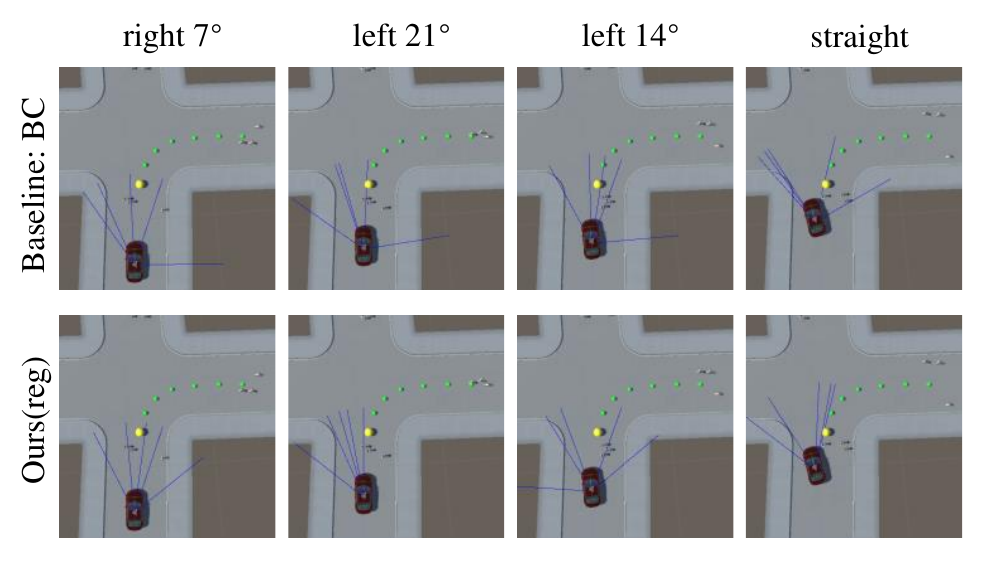}
    \caption{The top-5 important radars on driving (blue lines).
    The caption of each figure shows the corresponding wheel action. For example, ``right $7^{\circ}$'' represents a 7 degrees rotation to the right. We highlight the top-5 important ones from 64 radars covering all the forward directions.
    } 
    \label{autopilot}
\end{figure}

We also test our algorithm on an autopilot simulation environment.
In this task, the agent needs to drive the car to the destination safely, meanwhile, pedestrians are walking randomly on the road.
The action sets include the steering wheel and the throttle, and each has 11 discrete actions. Different from previous experiments, the neural network policy is trained by PPO algorithm. 
We calculate the advantage by Equation (\ref{softq}) and construct trees from 200, 000 samples. The observation includes 64 laser beams up front cover 180 degrees.  
We sample a trajectory and mark out the top-5 important radar bearings per frame. Figure \ref{autopilot} shows sample states where our approach produces a more meaningful saliency feature than the baseline approach for the autopilot agent. Even though the two policies both focus on the pedestrian or curb locations, the traditional decision tree policy doesn't focus on orientation while ours focuses more on the obstacles ahead. 
The whole process is shown in Appendix C.

\section{Conclusion and Future Work}
We have applied the model distillation on reinforcement learning to generate a tree-based policy. Considering the particularity of RL, we described a decision tree method for mimicking an existed neural network policy, which maximizes the cumulative rewards instead of classification accuracy. Experiments on a series of tasks show the improvements in both performance and consistency with the original policy, which verify that it is an effective way to employ the information outside the demonstrated action, e.g., Q, V, and demonstrated action distribution. Finally, our approach can be applied conveniently to even real games and industrial simulators, which may help provide interpretability and security for the automatic system. 

\newpage
\bibliographystyle{named}
\bibliography{ijcai21}

\begin{thebibliography}{}

\bibitem[\protect\citeauthoryear{Bastani \bgroup \em et al.\egroup
  }{2016}]{bastani2016measuring}
Osbert Bastani, Yani Ioannou, Leonidas Lampropoulos, Dimitrios Vytiniotis,
  Aditya Nori, and Antonio Criminisi.
\newblock Measuring neural net robustness with constraints.
\newblock In {\em NeurIPS'16}, 2016.

\bibitem[\protect\citeauthoryear{Bastani \bgroup \em et al.\egroup
  }{2018}]{bastani2018verifiable}
Osbert Bastani, Yewen Pu, and Armando Solar-Lezama.
\newblock Verifiable reinforcement learning via policy extraction.
\newblock In {\em NeurIPS’18}, 2018.

\bibitem[\protect\citeauthoryear{Chang \bgroup \em et al.\egroup
  }{2019}]{chang2018explaining}
Chun-Hao Chang, Elliot Creager, Anna Goldenberg, and David Duvenaud.
\newblock Explaining image classifiers by counterfactual generation.
\newblock In {\em ICLR'19}, 2019.

\bibitem[\protect\citeauthoryear{Che \bgroup \em et al.\egroup
  }{2015}]{healthcare}
Zhengping Che, Sanjay Purushotham, Robinder Khemani, and Yan Liu.
\newblock Distilling knowledge from deep networks with applications to
  healthcare domain.
\newblock {\em arXiv preprint arXiv:1512.03542}, 2015.

\bibitem[\protect\citeauthoryear{Chen \bgroup \em et al.\egroup
  }{2019}]{verif-tree}
Hongge Chen, Huan Zhang, Si~Si, Yang Li, Duane Boning, and Cho-Jui Hsieh.
\newblock Robustness verification of tree-based models.
\newblock In {\em NeurIPS'19}, 2019.

\bibitem[\protect\citeauthoryear{Coppens \bgroup \em et al.\egroup
  }{2019}]{coppens2019distilling}
Youri Coppens, Kyriakos Efthymiadis, Tom Lenaerts, Ann Now{\'e}, T~Miller,
  R~Weber, and D~Magazzeni.
\newblock Distilling deep reinforcement learning policies in soft decision
  trees.
\newblock In {\em IJCAI'19 Workshop on Explainable Artificial Intelligence},
  2019.

\bibitem[\protect\citeauthoryear{Czarnecki \bgroup \em et al.\egroup
  }{2019}]{dis_policy_distillation}
W.~Czarnecki, Razvan Pascanu, Simon Osindero, Siddhant~M. Jayakumar,
  G.~Swirszcz, and Max Jaderberg.
\newblock Distilling policy distillation.
\newblock In {\em AISTATS’19}, 2019.

\bibitem[\protect\citeauthoryear{Doshi-Velez and Kim}{2017}]{doshi2017towards}
Finale Doshi-Velez and Been Kim.
\newblock Towards a rigorous science of interpretable machine learning.
\newblock {\em arXiv preprint arXiv:1702.08608}, 2017.

\bibitem[\protect\citeauthoryear{Ernst \bgroup \em et al.\egroup
  }{2005}]{ernst2005tree}
Damien Ernst, Pierre Geurts, and Louis Wehenkel.
\newblock Tree-based batch mode reinforcement learning.
\newblock {\em J Mach Learn Res}, 2005.

\bibitem[\protect\citeauthoryear{Frosst and
  Hinton}{2017}]{frosst2017distilling}
Nicholas Frosst and Geoffrey Hinton.
\newblock Distilling a neural network into a soft decision tree.
\newblock {\em arXiv preprint arXiv:1711.09784}, 2017.

\bibitem[\protect\citeauthoryear{Goodfellow \bgroup \em et al.\egroup
  }{2016}]{Deep_learning}
Ian Goodfellow, Yoshua Bengio, Aaron Courville, and Yoshua Bengio.
\newblock {\em Deep Learning}.
\newblock MIT press Cambridge, 2016.

\bibitem[\protect\citeauthoryear{Gretton \bgroup \em et al.\egroup
  }{2012}]{gretton2012kernel}
Arthur Gretton, Karsten~M Borgwardt, Malte~J Rasch, Bernhard Sch{\"o}lkopf, and
  Alexander Smola.
\newblock A kernel two-sample test.
\newblock {\em J Mach Learn Res}, 2012.

\bibitem[\protect\citeauthoryear{Haarnoja \bgroup \em et al.\egroup
  }{2017}]{haarnoja2017reinforcement}
Tuomas Haarnoja, Haoran Tang, Pieter Abbeel, and Sergey Levine.
\newblock Reinforcement learning with deep energy-based policies.
\newblock In {\em ICML'17}, 2017.

\bibitem[\protect\citeauthoryear{Horgan \bgroup \em et al.\egroup
  }{2018}]{horgan2018distributed}
Dan Horgan, John Quan, David Budden, Gabriel Barth-Maron, Matteo Hessel, Hado
  van Hasselt, and David Silver.
\newblock Distributed prioritized experience replay.
\newblock In {\em ICLR'18}, 2018.

\bibitem[\protect\citeauthoryear{Levine}{2018}]{levine2018reinforcement}
Sergey Levine.
\newblock Reinforcement learning and control as probabilistic inference:
  Tutorial and review.
\newblock {\em arXiv preprint arXiv:1805.00909}, 2018.

\bibitem[\protect\citeauthoryear{Liu \bgroup \em et al.\egroup
  }{2018}]{liu2018toward}
Guiliang Liu, Oliver Schulte, Wang Zhu, and Qingcan Li.
\newblock Toward interpretable deep reinforcement learning with linear model
  u-trees.
\newblock In {\em ECMLPKDD'18}, 2018.

\bibitem[\protect\citeauthoryear{Molnar \bgroup \em et al.\egroup
  }{2020}]{molnar2019interpretable}
Christoph Molnar, Giuseppe Casalicchio, and Bernd Bischl.
\newblock Interpretable machine learning--a brief history, state-of-the-art and
  challenges.
\newblock {\em arXiv preprint arXiv:2010.09337}, 2020.

\bibitem[\protect\citeauthoryear{Neftci and
  Averbeck}{2019}]{neftci2019reinforcement}
Emre~O. Neftci and Bruno~B. Averbeck.
\newblock Reinforcement learning in artificial and biological systems.
\newblock {\em Nat. Mach. Intell.}, 2019.

\bibitem[\protect\citeauthoryear{O'Kelly \bgroup \em et al.\egroup
  }{2018}]{o2018scalable}
Matthew O'Kelly, Aman Sinha, Hongseok Namkoong, Russ Tedrake, and John~C Duchi.
\newblock Scalable end-to-end autonomous vehicle testing via rare-event
  simulation.
\newblock In {\em NeurIPS'18}, 2018.

\bibitem[\protect\citeauthoryear{Pedregosa \bgroup \em et al.\egroup
  }{2011}]{scikit-learn}
F.~Pedregosa, G.~Varoquaux, A.~Gramfort, V.~Michel, B.~Thirion, O.~Grisel,
  M.~Blondel, P.~Prettenhofer, R.~Weiss, V.~Dubourg, J.~Vanderplas, A.~Passos,
  D.~Cournapeau, M.~Brucher, M.~Perrot, and E.~Duchesnay.
\newblock Scikit-learn: Machine learning in {P}ython.
\newblock {\em J Mach Learn Res}, 2011.

\bibitem[\protect\citeauthoryear{Popova \bgroup \em et al.\egroup
  }{2018}]{popova2018deep}
Mariya Popova, Olexandr Isayev, and Alexander Tropsha.
\newblock Deep reinforcement learning for de novo drug design.
\newblock {\em Sci. Adv.}, 2018.

\bibitem[\protect\citeauthoryear{Reddy \bgroup \em et al.\egroup
  }{2020}]{reddy2019sqil}
Siddharth Reddy, Anca~D. Dragan, and Sergey Levine.
\newblock Sqil: Imitation learning via reinforcement learning with sparse
  rewards.
\newblock In {\em ICLR'20}, 2020.

\bibitem[\protect\citeauthoryear{Ross and Bagnell}{2010}]{ross2010efficient}
Stephane Ross and Drew Bagnell.
\newblock Efficient reductions for imitation learning.
\newblock In {\em AISTATS‘10}, 2010.

\bibitem[\protect\citeauthoryear{Ross \bgroup \em et al.\egroup
  }{2011}]{ross2011reduction}
Stephane Ross, Geoffrey Gordon, and Drew Bagnell.
\newblock A reduction of imitation learning and structured prediction to
  no-regret online learning.
\newblock In {\em AISTATS'11}, 2011.

\bibitem[\protect\citeauthoryear{Rusu \bgroup \em et al.\egroup
  }{2015}]{policy_distillation}
Andrei~A Rusu, Sergio~Gomez Colmenarejo, Caglar Gulcehre, Guillaume Desjardins,
  James Kirkpatrick, Razvan Pascanu, Volodymyr Mnih, Koray Kavukcuoglu, and
  Raia Hadsell.
\newblock Policy distillation.
\newblock {\em arXiv preprint arXiv:1511.06295}, 2015.

\bibitem[\protect\citeauthoryear{Schulman \bgroup \em et al.\egroup
  }{2015}]{schulman2015trust}
John Schulman, Sergey Levine, Pieter Abbeel, Michael Jordan, and Philipp
  Moritz.
\newblock Trust region policy optimization.
\newblock In {\em ICML'15}, 2015.

\bibitem[\protect\citeauthoryear{Schwab and Karlen}{2019}]{NIPS2019_9211}
Patrick Schwab and Walter Karlen.
\newblock Cxplain: Causal explanations for model interpretation under
  uncertainty.
\newblock In {\em NeurIPS'19}, 2019.

\bibitem[\protect\citeauthoryear{Singla \bgroup \em et al.\egroup
  }{2020}]{singla2019explanation}
Sumedha Singla, Brian Pollack, Junxiang Chen, and Kayhan Batmanghelich.
\newblock Explanation by progressive exaggeration.
\newblock In {\em ICLR'20}, 2020.

\bibitem[\protect\citeauthoryear{Tang \bgroup \em et al.\egroup
  }{2019}]{tang2019deep}
Xiaocheng Tang, Zhiwei~(Tony) Qin, Fan Zhang, Zhaodong Wang, Zhe Xu, Yintai Ma,
  Hongtu Zhu, and Jieping Ye.
\newblock A deep value-network based approach for multi-driver order
  dispatching.
\newblock In {\em KDD'19}, 2019.

\bibitem[\protect\citeauthoryear{Zhang \bgroup \em et al.\egroup
  }{2019}]{zhang2019interpreting}
Quanshi Zhang, Yu~Yang, Haotian Ma, and Ying~Nian Wu.
\newblock Interpreting cnns via decision trees.
\newblock In {\em CVPR'19}, 2019.

\end{thebibliography}

\end{document}